\documentclass[letterpaper]{article} % DO NOT CHANGE THIS
\usepackage{aaai25}  % DO NOT CHANGE THIS
\usepackage{times}  % DO NOT CHANGE THIS
\usepackage{helvet}  % DO NOT CHANGE THIS
\usepackage{courier}  % DO NOT CHANGE THIS
\usepackage[hyphens]{url}  % DO NOT CHANGE THIS
\usepackage{graphicx} % DO NOT CHANGE THIS
\usepackage{natbib}  % DO NOT CHANGE THIS AND DO NOT ADD ANY OPTIONS TO IT
\usepackage{caption} % DO NOT CHANGE THIS AND DO NOT ADD ANY OPTIONS TO IT
\usepackage{algorithm}
\usepackage{algorithmic}

\usepackage{newfloat}
\usepackage{listings}
\DeclareCaptionStyle{ruled}{labelfont=normalfont,labelsep=colon,strut=off} % DO NOT CHANGE THIS
\floatstyle{ruled}
\newfloat{listing}{tb}{lst}{}
\floatname{listing}{Listing}
\def\mc{\mathcal}

\usepackage{multirow}
\usepackage{graphicx}
\usepackage{caption}
\usepackage{booktabs}
\usepackage{xcolor}
\usepackage{amssymb}

\usepackage{amsmath}

\nocopyright
\title{Rebalancing Multi-Label Class-Incremental Learning}
\author{
Kaile Du\textsuperscript{\rm 1},
Yifan Zhou\textsuperscript{\rm 1},
Fan Lyu\textsuperscript{\rm 2},
Yuyang Li\textsuperscript{\rm 1},
Junzhou Xie\textsuperscript{\rm 1},\\
Yixi Shen\textsuperscript{\rm 3},
Fuyuan Hu\textsuperscript{\rm 3},
Guangcan Liu\textsuperscript{\rm 1} \thanks{Corresponding author.}
}
\affiliations{
    \textsuperscript{\rm 1}School of Automation, Southeast University, China
\\
    \textsuperscript{\rm 2}NLPR, MAIS, CASIA, China\\
    \textsuperscript{\rm 3}School of Electronic and Information Engineering, Suzhou University of Science and Technology, China\\
    
}

\usepackage{bibentry}
\begin{document}

\maketitle

\begin{abstract}
Multi-label class-incremental learning (MLCIL) is essential for real-world multi-label applications, allowing models to learn new labels while retaining previously learned knowledge continuously. 
However, recent MLCIL approaches can only achieve suboptimal performance due to the oversight of the positive-negative imbalance problem, which manifests at both the label and loss levels because of the task-level partial label issue. 
The imbalance at the label level arises from the substantial absence of negative labels, while the imbalance at the loss level stems from the asymmetric contributions of the positive and negative loss parts to the optimization. 
To address the issue above, we propose a {Reb}alance framework for both the {L}oss and {L}abel levels (RebLL), which integrates two key modules: asymmetric knowledge distillation (AKD) and online relabeling (OR). AKD is proposed to rebalance at the loss level by emphasizing the negative label learning in classification loss and down-weighting the contribution of overconfident predictions in distillation loss.
OR is designed for label rebalance, which restores the original class distribution in memory by online relabeling the missing classes.
Our comprehensive experiments on the PASCAL VOC and MS-COCO datasets demonstrate that this rebalancing strategy significantly improves performance, achieving new state-of-the-art results even with a vanilla CNN backbone.

\end{abstract}
\section{Introduction}

\label{sec:intro}
\begin{figure}
    \centering
    \includegraphics[width=\linewidth]{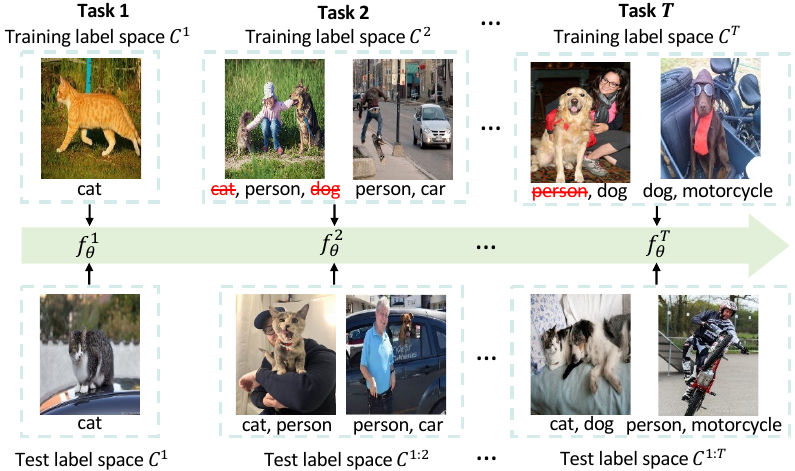}
    \caption{A diagram of multi-label class-incremental learning. Labels are trained and tested across tasks from Task $1$ to Task $T$. Missing labels are highlighted in red.  The training label space is task-specific, while the testing label space progressively expands with the addition of each new task.} 
    \label{fig:figure1}
\end{figure}

Class-incremental learning (CIL) \cite{buzzega2020dark,douillard2020podnet,wang2023comprehensive,lyu2024variational} continuously learn new classes without the need to retrain on the entire dataset. 
 However, as the model learns new classes, it may overwrite the knowledge of previous tasks, resulting in a significant decline in performance on older classes, a phenomenon known as catastrophic forgetting
\cite{mccloskey1989catastrophic}. While numerous methods have been developed for single-label class-incremental learning (SLCIL) \cite{douillard2020podnet,wang2022learning,wang2023comprehensive,lyu2023measuring}, there has been limited exploration into the more practical domain of multi-label class-incremental learning (MLCIL) \cite{dong2023knowledge,du2024confidence}. 
% Numerous methods have been developed for single-label class-incremental learning (SLCIL) \cite{buzzega2020dark}, but only limited studies explored the more practical multi-label class-incremental learning (MLCIL) \cite{dong2023knowledge,du2024confidence}.
In MLCIL, training images for new classes are partially labeled to minimize the cost of manual annotation. Specifically, only the classes for the new task are labeled, while the past and future ones are not annotated, a phenomenon called task-level partial label (PL)  \cite{du2024confidence} or category-incomplete  \cite{dong2023knowledge} issue.
For example, as shown in Figure \ref{fig:figure1}, there is a training image of Task $2$ containing the labels ``cat'', ``person'' and ``dog'', only the labels for the current task (``person'') are annotated, while the labels for past (``cat'') and future (``dog'') tasks remain unannotated. %\du{This is a common description of PL in previous works \cite{10221710,dong2023knowledge,du2024confidence}, where it is evident that they focus more on the presence or absence of positive labels.}

% In multi-label learning, data is likely to have more negative labels than positive labels \cite{ridnik2021asymmetric,zhou2022acknowledging,xie2024class}. 
% Due to the PL setting, the missing labels are default viewed as negative ones.
However, most recent MLCIL methods focus on designing new network architectures for MLCIL to fit the PL setting, ignoring the positive-negative imbalance issue.
{AGCN \cite{10221710} utilizes pseudo-labels to construct statistical label relationships within the Graph Convolutional Network (GCN). 
KRT \cite{dong2023knowledge} is the knowledge restoration and transfer framework based on a cross-attention mechanism (CAM).  
CSC \cite{du2024confidence} leverages a learnable GCN to calibrate label relationships.
}
In fact, due to the larger number of negative labels compared to positive labels in multi-label samples \cite{ridnik2021asymmetric,zhou2022acknowledging,xie2024class}, missing labels for both past and future tasks are prone to be labeled negative in MLCIL. A training example of task $T$ is shown in Figure  \ref{fig:figure1}, it only lacks one positive label ``person'' (in red), while also missing many negative labels (``cat...car'') from other tasks.
% where the task-specific labels for Task $T$ are ``dog'' and ``motorcycle''. 
% When training on a sample that contains both ``dog'' and ``motorcycle'', it only has two positive labels and does not include other labels within the image. 
% This situation poses the risk of collapsing to a trivial solution, i.e., the model tends to predict labels as positive ones \cite{zhou2022acknowledging,liu2023revisiting}. 
% The substantial absence of negative labels leads to asymmetric learning of positive and negative labels by the model.
The substantial absence of negative labels (more than 9 times that of positive) results in a positive-negative imbalance at the label level. In some cases, the model may only learn the positive labels. 
\textit{Learning on such imbalance-labeled data poses the risk of collapsing to a trivial solution}, i.e., the model tends to predict each label as positive \cite{zhou2022acknowledging,liu2023revisiting}, thus leading to numerous false positive errors during loss optimization. This positive-negative imbalance manifests at the loss level due to the asymmetric contributions of the positive and negative loss parts to the optimization objectives.
% This asymmetric absence of positive and negative labels due to the PL issue leads to an imbalance between them.
% Learning on such imbalance-labeled data poses the risk of collapsing to a trivial solution, i.e., the model tends to predict each label as positive \cite{zhou2022acknowledging,liu2023revisiting}, thus leading to numerous false positive errors.
% \du{This confusion between positive and negative labels worsens with class-incremental learning, thereby exacerbating forgetting.}

{In this paper, we explore tackling the positive-negative imbalance issue in MLCIL from two key aspects: loss level and label level.
}
The positive-negative imbalance arises at the \textit{loss level} of recent MLCIL methods \cite{10221710,du2024confidence}. Specifically, they utilize binary cross-entropy (BCE) loss and binary knowledge distillation (KD) loss for classification and mitigating forgetting. The loss imbalance during the optimization process fundamentally stems from the equal treatment of imbalanced positive and negative loss components by the BCE and KD losses. 
As a result, the contribution of positive labels to the loss is significantly greater than that of negative labels. This situation is suboptimal, as it leads to the accumulation of more loss gradients from positive labels due to the positive-negative imbalance \cite{ridnik2021asymmetric}, resulting in insufficient learning of negative labels at the loss level.
Due to the substantial absence of negative labels, the positive-negative imbalance also occurs at the \textit{label level}. We aim to address this issue using a replay-based approach.
In our further attempts to mitigate catastrophic forgetting, we observe that replay methods \cite{rolnick2019experience,douillard2020podnet,buzzega2020dark}, commonly used in single-label scenarios, exacerbate the label-level imbalance due to memory sampling of imbalance-labeled data. This exacerbation leads to an increase in false positive errors. Consequently, this label-level imbalance causes these methods to suboptimally address catastrophic forgetting.
To address the issues above, we first propose asymmetric knowledge distillation (AKD) for loss rebalance. AKD emphasizes the learning of negative labels in the classification loss for the current model and down-weights the contribution of overconfident old task predictions in the KD process. 
% based on classical distillation and replay-based anti-forgetting methods \cite{li2017learning,rolnick2019experience,buzzega2020dark,douillard2020podnet}.
% This framework consists of two components: asymmetric knowledge distillation (AKD) and online relabeling (OR). 
% 1) In traditional bce applied to recent MLCIL method, positive and negative labels are treated equally, leading to suboptimal results in the context of positive-negative imbalance in MLCIL,
% Compared to the traditional bce loss applied to recent MLCIL method, our proposed AKD loss operates differently on positive and negative labels. 
% The AKD loss rebalances positive and negative labels at optimization objective level by emphasizing the learning of negative labels in the classification loss for the current model and down-weighting the contribution of overconfident old task predictions based on KD loss. 
% However, previous MLCIL work has overlooked this issue, leading to their suboptimal performance.
Then, we design the online relabeling (OR) strategy for {label rebalance} by continuously restoring the original class distribution in memory.
Specifically, we relabel the missing old labels of the memory samples using the past model, while the missing new labels in memory are labeled with the trained current model.
\textbf{Reb}alancing at both the \textbf{L}oss and \textbf{L}abel levels, forming our \textbf{RebLL} framework.
 Our framework can mitigate the positive-negative imbalance and enhance MLCIL performance. 
 % the effectiveness of anti-forgetting methods in MLCIL by balancing the learning of positive and negative labels.
% Under the PL setting, our contributions are summarized below:
% Our rebalancing framework can enhance the effectiveness of anti-forgetting methods in MLCIL by balancing the learning of positive and negative labels.
Under the PL setting, our contributions are summarized below:

\begin{itemize}
\item {We are the first to identify the inherent positive-negative imbalance in MLCIL under the PL setting. This imbalance, occurring at both the loss and label levels, impedes the effective learning of MLCIL.
% This imbalance can lead to numerous false positive errors, hindering the model's ability to perform MLCIL effectively.
}
\item {We propose a RebLL framework to suppress the positive-negative imbalance, consisting of AKD and OR components. AKD is designed for loss rebalance, while OR targets label rebalance. RebLL can effectively mitigate forgetting and enhance MLCIL performance by suppressing the positive-negative imbalance.} 

% AKD is designed to emphasize negative sample learning,  while down-weighting the contribution of overconfident old task predictions. OR strategy is introduced to continuously restore the original label distribution of memory samples by online relabeling the missing classes. Thus, it achieves a rebalancing of positive and negative label learning, effectively mitigating catastrophic forgetting and enhancing MLCIL performance.}
    % \item {An asymmetric knowledge distillation loss is designed to down-weight overconfident pseudo-labels from the past model and emphasize the contribution of negative samples for the current model, thereby facilitating knowledge distillation to more effectively mitigate catastrophic forgetting by reducing the false positive rate.}
    % \item {An online relabeling strategy is proposed to continuously restore the original label distribution of memory samples by online completion of the missing labels, thereby alleviating the impact of partial labeling on replay method.}
    \item {
    Extensive experiments conducted across multiple MLCIL scenarios using the PASCAL VOC and MS-COCO  demonstrate that RebLL achieves new SOTA results, regardless of whether we use a vanilla CNN or a more powerful network architecture as the backbone.
    }
\end{itemize}
\section{Related Work}
\subsection{Single-Label Class-Incremental Learning} 

Significant progress has been made in single-label class-incremental learning in recent years. Regularization-based methods \cite{li2017learning,lyu2020multi,sun2022exploring,zhao2023few,mohamed2023d3former} impose constraints on the loss function to preserve old knowledge from being overwritten by new one. 
% These methods do not require the storage of data from old tasks; instead, they leverage the knowledge of previous tasks to guide the learning of new ones. 
For instance, EWC \cite{kirkpatrick2017overcoming} employs the Fisher matrix to retain the critical parameters associated with earlier tasks. LwF \cite{li2017learning} employs knowledge distillation to transfer old knowledge to new tasks.
% , thereby mitigating catastrophic forgetting.
Replay-based methods \cite{rolnick2019experience,de2021continual,ye2021lifelong,cha2023rebalancing,luo2023class} enhance the learning of new data by incorporating a portion of old data, thereby reinforcing previously acquired knowledge. 
% While this approach effectively mitigates the issue of catastrophic forgetting, it incurs additional storage and computational costs. 
Experience replay \cite{rolnick2019experience} randomly samples a subset of sample from the old task dataset to serve as memory data. DER++ \cite{buzzega2020dark} combines replay with knowledge distillation and regularization techniques.
% PODNet \cite{douillard2020podnet} distills knowledge from the intermediate and final layers.
Architecture-based approaches assign distinct parameters to each task to save the corresponding knowledge. Some architecture-based methods decompose the model into task-sharing and task-specific components \cite{douillard2022dytox,wu2021incremental}. Furthermore, L2P \cite{wang2022learning} introduces prompt-driven learning into the domain of class-incremental learning, yielding encouraging results by leveraging a pre-trained ViT model \cite{dosovitskiy2020image}.
\begin{figure*}
    \centering
    \includegraphics[width=0.85\linewidth]{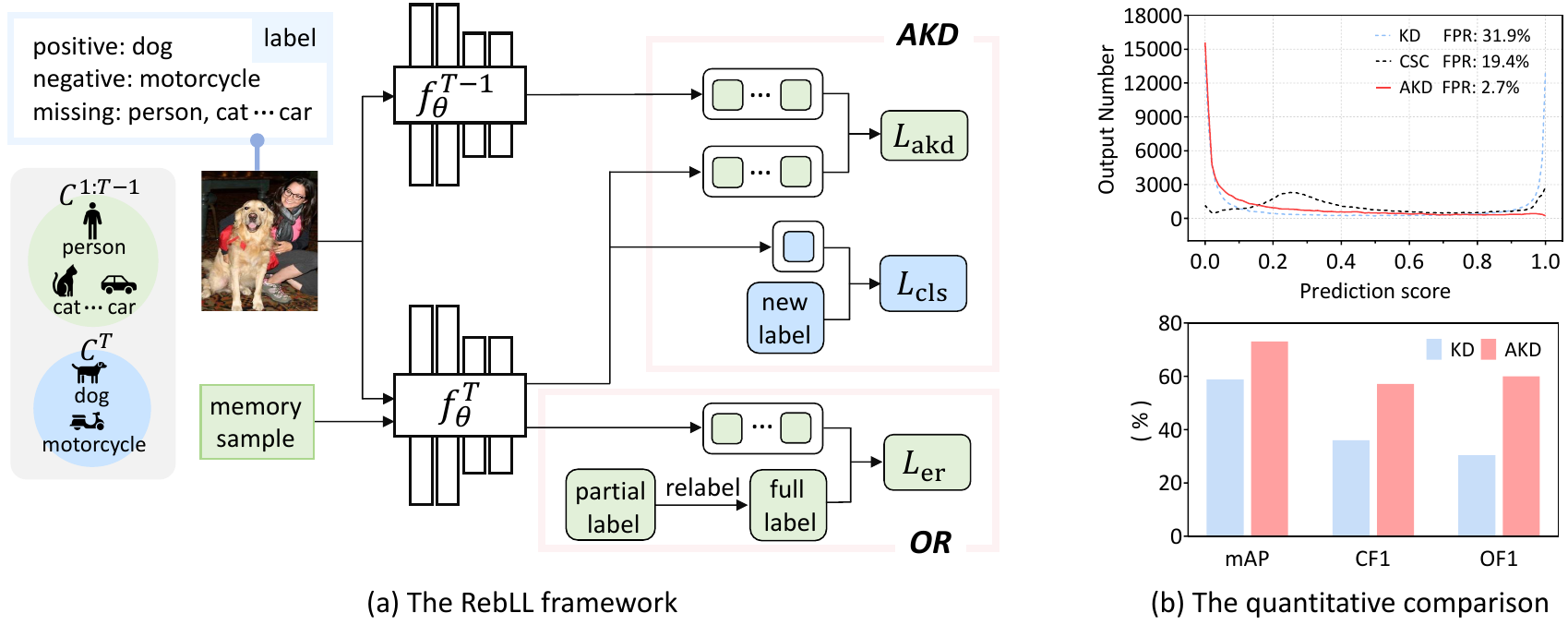}
    \caption{
    (a) The RebLL framework consists of two modules: AKD and OR. 
    %Given a training image containing labels ``dog'' and ``person'', its ground truth (GT) only includes task-specific positive label ``dog'' and  negative label ``motorcycle''.
    %Given a training image containing ``dog'' and ``person'', since the ``person'' has already been trained in a previous task, the ground truth (GT) for this image only includes positive label ``dog'' and  negative label ``motorcycle''. 
    % A training image is fed into both  old ($f^{T-1}_{\theta}$) and new ($f^{T}_{\theta}$) models, where the output old task predictions  (in green) are used to compute \(L_{\text{akd}}\). The new task predictions (in blue) are compared with the new GT (n-GT) to compute \(L_{\text{cls}}\). The partially labeled GT (p-GT) of memory sample is relabeled to form the complete old GT (o-GT), which is then used in conjunction with the predictions to compute \(L_{\text{er}}\). 
    In AKD, a training image is fed into both old ($f^{T-1}_{\theta}$) and new ($f^{T}_{\theta}$) models, where the output old task predictions  (in green) are used to compute \(L_{\text{akd}}\). The new task predictions (in blue) are compared with the new label to compute \(L_{\text{cls}}\).
    In OR, the partially labeled memory sample is relabeled to form the full label, which is then used in conjunction with the predictions to compute \(L_{\text{er}}\). 
     % The partially labeled GT  of memory sample is relabeled to form the full GT, which is then used in conjunction with the predictions to compute \(L_{\text{er}}\). 
    % (a) The qualitative description of asymmetric knowledge distillation. The asymmetric knowledge distillation (AKD) consists of $L_{akd}$ and $L_{cls}$. When a sample with the labels "dog" and "person" is trained, the last task model $f^{T-1}_{\theta}$ and current task model $f^{T}_{\theta}$  produce the corresponding prediction scores. Compared to classical KD, AKD down-weights the overconfident pseudo-labels (``cat'' and ``car'') through $L_{akd}$, while $L_{cls}$ emphasizes the role of the negative label ``motorcycle''.
    (b) The quantitative comparison between KD, CSC and AKD after training on the final task in \{B4-C2\} of VOC 2007.
    } 
    \label{fig:figure2}
\end{figure*}

\subsection{Multi-Label Class-Incremental Learning
}
Multi-label class-incremental learning is an emerging field.
PRS \cite{kim2020imbalanced} and OCDM \cite{liang2022optimizing} introduce sampling approaches designed for replay methods, aimed at alleviating the effects of multi-label long-tail class distributions. The methods above differ significantly from recent MLCIL approaches \cite{9859622, dong2023knowledge,10221710,du2024confidence} in both benchmarks and settings. AGCN \cite{10221710} utilizes pseudo-labels to construct statistical label relationships within the GCN to enhance MLCIL performance. Meanwhile, KRT \cite{dong2023knowledge} is the knowledge restoration and transfer framework based on a cross-attention mechanism to address the category-incomplete issue.  
% However, these methods overlook the significant prevalence of false positive (F-P) errors arising from overly confident output distributions, indicating a deficiency in model calibration. 
CSC \cite{du2024confidence} leverages a learnable GCN to calibrate label relationships and employs maximum entropy to adjust overconfident predictions.
Notably, CSC is the SOTA method. It can be observed that AGCN, KRT, and CSC all utilize powerful network architectures as backbones, while our approach, by rebalancing the inherent positive-negative imbalance in MLCIL, allows us to surpass them using only a vanilla CNN.

\section{Method}
\subsection{Preliminary}
\subsubsection{MLCIL.} 

Following previous works~\cite{10221710,dong2023knowledge,du2024confidence}, given \textit{T} multi-label learning tasks, and the corresponding training sets for these tasks are denoted as $\{\mc{D}^{1}_\text{trn},\cdots,\mc{D}^{T}_\text{trn}\}$, and the testing sets are represented as \(\{\mc{D}^{1}_\text{tst},\cdots,\mc{D}^{T}_\text{tst}\}\). 
For each incremental state \( t \), the training dataset is defined as $\mc{D}^{t}_\text{trn}$. 
% For each incremental state \( t \), the training dataset is defined as \(\mc{D}^{t}_\text{trn}=\{(x^t_i,y^t_i)\}\).
% where \( x^t_i \) denotes the \( i \)-th training sample, and \( y^t_i \in \mc{Y}^t_\text{trn} \) represents the corresponding ground-truth labels. 
$\mc{C}^t$ denotes the current task-specific class set, while the past class collection is denoted by $\mc{C}^{1:t-1}=\bigcup_{i=1}^{t-1}\mc{C}^i$. Moreover, $\mc{C}^{1:t} = \mc{C}^{1:t-1}\cup\mc{C}^{t}$, and it holds that $\mc{C}^{1:t-1}\cap\mc{C}^t=\varnothing$. The goal of MLCIL is for the model to recognize all previously learned labels in a multi-label image after sequentially training task-specific labels. Hence, for the \( t \)-th task, the training label space \(\mc{Y}^t_\text{trn}\) = \(\mc{C}^t\), the testing label space  \(\mc{Y}^t_\text{tst}=\mc{C}^{1:t}\).
 
As shown in Figure \ref{fig:figure2} (a), similar to distillation-based MLCIL methods \cite{10221710,dong2023knowledge,du2024confidence}, we maintain a fixed old model $f^{T-1}_{\theta}$ and a trainable new model $f^{T}_{\theta}$. 
% Our rebalancing framework, composed of the AKD and OR modules, is designed to mitigate this positive-negative imbalance and enhance the MLCIL performance. 
 % When operating on these imbalance-labeled data with traditional CIL methods, the positive-negative imbalance can arise at both the loss and label levels.
 In MLCIL, the inherent positive-negative imbalance manifests at loss and label levels. To address this issue, we propose the RebLL framework, which includes asymmetric knowledge distillation (AKD) for loss rebalance and online relabeling (OR) for label rebalance.
 % We propose a RebLL framework to suppress the positive-negative imbalance, consisting of AKD and OR. AKD is proposed for loss rebalance, while OR is designed for label rebalance.
% This positive-negative imbalance can be viewed as occurring at both the optimization objective and memory sample levels.
% \subsubsection{Loss function in existing methods.}
% \subsubsection{Imbalance in optimization objective.}
\subsubsection{Imbalance in Loss.}
Recent MLCIL~\cite{10221710,du2024confidence} methods utilize BCE and KD for classification and mitigating forgetting.
The BCE loss decouples the multi-label classification tasks into multiple binary issues,  formulated as:
% which : 
\begin{equation}\label{eq:bce}
L_{\text{bce}} = \sum_{c \in \mc{C}^{t}} -y_c^t \log(\hat{y}_c^t) - (1-y_c^t) \log(1-\hat{y}_c^t),
\end{equation}
where $\hat{y}_c^t$ and $y_c^t$ represents the predcitions and ground-truth for class $c$ and task $t$.
KD loss uses the last task prediction  $\hat{y}_c^{t-1}$  as supervisory information to guide the learning of current task to suppress catastrophic forgetting, formulated as:
\begin{equation}\label{eq:kd}
L_{\text{kd}} = \sum_{c \in \mc{C}^{1:t-1}} -\hat{y}_c^{t-1} \log(\hat{y}_c^{t}) - (1-\hat{y}_c^{t-1}) \log(1-\hat{y}_c^t).
\end{equation}
To clarify how positive and negative labels are operated in $L_{\text{bce}}$, a general form of Eq.~\eqref{eq:bce}  can be given by:
$L_{\text{bce}} = \sum -y_c^t L_+ - (1-y_c^t) L_-$
, $L_+$ and $L_-$ can be regarded positive and negative loss parts,  $L_{\text{kd}}$  follows the same format. 
Both $L_{\text{bce}}$ and $L_{\text{kd}}$ treat positive and negative loss parts equivalently, making it unsuitable for MLCIL  where negative labels are insufficiently learned using the imbalance-labeled data. 
This approach is suboptimal because it results in the accumulation of excessive loss gradients from positive samples under positive-negative imbalance \cite{ridnik2021asymmetric}. 
% {It is suboptimal as the accumulation of more loss gradients from positive samples under the positive-negative imbalance \cite{ridnik2021asymmetric}.} 
At the loss level,  as there is an asymmetric learning of positive and negative labels in MLCIL, the contribution of the negative loss part should be emphasized, while the positive loss part should be down-weighted.
% The logits $p_c^t$ are activated by the sigmoid function $\sigma(p_c^t)$ to get the corresponding prediction $\hat{y}_c^t$.
% The recent MLCIL methods employ KD techniques composed of Eq. (\ref{eq:bce}) and Eq. (\ref{eq:kd}) to mitigate forgetting. 
% A general form of Eq. (\ref{eq:kd})  can be given by: $L_{\text{kd}} = \sum -\hat{y}_c^{t-1} L_+ - (1-\hat{y}_c^{t-1}) L_-$, $L_+$ and $L_-$ can be regarded positive and negative loss parts,  $L_{\text{bce}}$  follows the same format. 

% under the asymmetric missing positive and negative labels in MLCIL. 
% It is not suitable for MLCIL under conditions of asymmetric missing positive and negative labels, which is suboptimal under the asymmetric missing positive and negative labels in MLCIL. 
%However, in MLCIL, the learning of negative labels is insufficient due to the positive-negative imbalance. Applying KD that treats positive and negative labels similarly yields suboptimal outcomes in mitigating forgetting. 
% Therefore, we should emphasize the learning of negative labels while down-weighting the positive label contribution. 
% Therefore, positive and negative loss parts asymmetrically at the optimization objective level. Based on this motivation, we propose the asymmetric knowledge distillation.

% \subsubsection{Imbalance in memory label.}
\subsubsection{Imbalance in Label.}
\begin{figure}
    \centering
    \includegraphics[width=\linewidth]{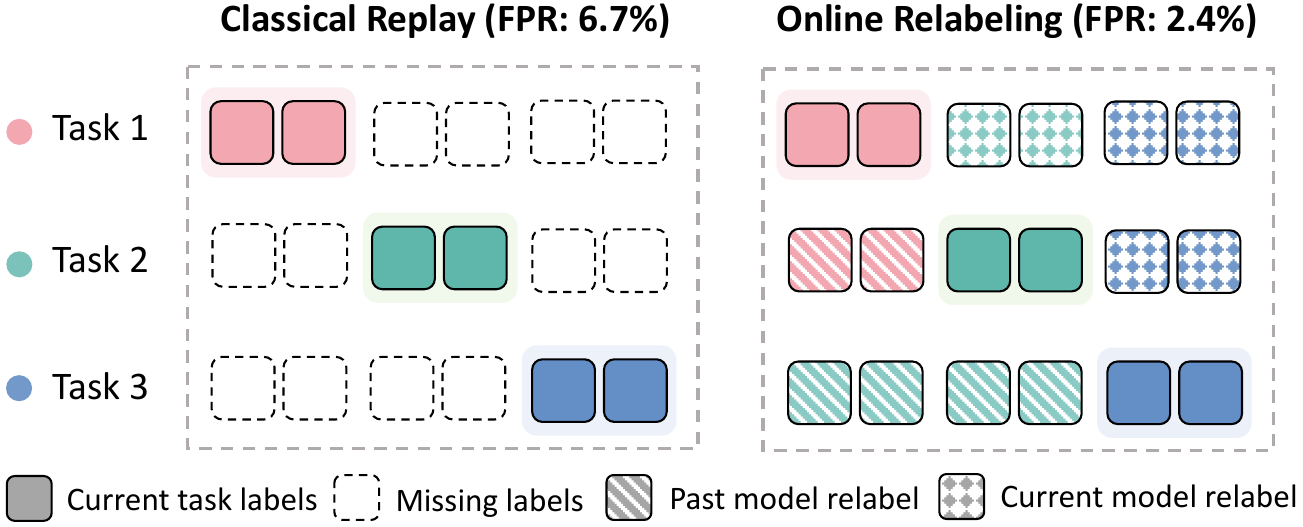}
    \caption{Online relabeling. For the label block matrix, missing new task labels above the main diagonal are relabeled using the trained current model, while the missing old task labels below the main diagonal are relabeled using the past model. This process reduces the FPR from 6.7\% to 2.4\%.
    % The FPR subsequently decreased from 6.7\% to 2.4\%.
    } 
    \label{fig:figure3}
\end{figure}

As shown in Figure \ref{fig:figure2} (a), given a multi-label image, its ground truth labels include a positive label ``dog'' and a negative label ``motorcycle''. In the PL setting, it is missing a positive label ``person'', while it lacks numerous negative labels, such as ``cat...car''. This asymmetric missing of positive and negative labels leads to a label-level imbalance in MLCIL. We aim to address this issue using a replay-based approach.
% As shown in Figure \ref{fig:figure3}, many future and past labels are missing in the memory data sampled by the classical replay method in SLCIL, which will exacerbate the imbalance in label during subsequent replay. When incorporating the classical replay method based on AKD, the false positive rate (FPR) increased from 2.7\% to 6.7\%. 
Figure \ref{fig:figure3} illustrates that many future and past labels are missing in the memory data sampled by the classical replay method in SLCIL. This exacerbates the imbalance in label during subsequent replay. When applying the classical replay method with AKD, the false positive rate (FPR) increased from 2.7\% to 6.7\%.
% , we find that the classical replay method samples imbalance-labeled examples from the current task's training data, which will exacerbate the positive-negative imbalance during subsequent replay, resulting in a higher false positive rate (FPR). 

% \du{Recently, most MLCIL methods, such as~\cite{10221710,dong2023knowledge,du2024confidence}, 
% leverage BCE loss and  KD loss to mitigate forgetting as follows.}

% which will result in suboptimal due to the inherent positive-negative imbalance in MLCIL. 
% As shown in Eq. (\ref{eq:bce}) and Eq. (\ref{eq:kd}), traditional binary cross-entropy applied in KD treats positive and negative labels equally, which will result in suboptimal due to the inherent positive-negative imbalance in MLCIL.
% Therefore, we propose the following asymmetric knowledge distillation.
% In traditional binary cross-entropy applied to KD, positive and negative labels are treated equally, leading to suboptimal results due to the inherent positive-negative imbalance in MLCIL. Our proposed AKD operates differently on positive-negative. It rebalances positive and negative labels by emphasizing the contribution of negative labels in the classification loss for the current model and down-weighting overconfident pseudo-labels from the past model in the KD loss.

% Due to the inherent positive-negative imbalance in MLCIL, directly applying classical knowledge distillation composed of Eq.~\eqref{eq:bce} and Eq.~\eqref{eq:kd} to MLCIL leads to suboptimal results. 
% \subsection{Asymmetric Knowledge Distillation} 
% \subsection{AKD for optimization objective rebalance} 
\subsection{AKD for Loss Rebalance} 
% In MLCIL, the PL issue means the model learns only the current task labels, which results in insufficient learning of negative labels. This insufficiency leads to the positive-negative imbalance. 
% In MLCIL, the imbalance makes  misclassification of many negative labels as positive, causing numerous false positive errors. 
% The classic KD method transfers the old task knowledge from the past task model to the current model to suppress catastrophic forgetting. When the trained old model tends to output positive predictions, this issue will also arise when transferring the old task knowledge to the new model. Thus, it cannot effectively mitigate catastrophic forgetting.
To rebalance the positive-negative at the loss level, we propose asymmetric knowledge distillation (AKD), which down-weights the contribution of overconfident predictions while emphasizing the learning of negative labels.
AKD consists of two components,  $L_{\text{cls}}$ and $L_{\text{akd}}$, which are enhancements of the BCE loss  $L_{\text{bce}}$ and KD loss $L_{\text{kd}}$.
% AKD consists of two parts $L_{\text{cls}}$ and $L_{\text{akd}}$, which are improvements over the BCE loss $L_{\text{bce}}$ and KD loss $L_{\text{kd}}$.

 % This leads to false positive errors, which are inherited by the current model $f^{t}_{\theta}$ through knowledge transfer. Our AKD addresses this issue by reducing such false positive predictions.

% Figure \ref{fig:figure2} (a) shows a qualitative comparison between AKD and KD, when training images contain only ``person'' and ``dog'' labels, in the classical KD method, the previous model $f^{t-1}_{\theta}$ still outputs high prediction scores for non-existent labels like ``car'' and ``cat''. This leads to false positive errors, which are inherited by the current model $f^{t}_{\theta}$ through knowledge transfer. Our AKD addresses this issue by reducing such false positive predictions.

Firstly, we modify  $L_\text{bce}$ in Eq.~\eqref{eq:bce} to $L_\text{cls}$.
We emphasize the learning of negative samples by down-weighting the positive ones in the classification loss for the current model:
\begin{equation}\label{eq:cls1}
L_{\text{cls}} = \sum_{c \in \mc{C}^{t}} -y_c^t L_+ - (1-y_c^t) \log(1-\hat{y}_c^t).
\end{equation}
% $L'_+$ is formulated as:
Inspired by focal loss \cite{lin2017focal}, $L_+$ is formulated as: 
\begin{equation}\label{eq:cls2}
L_+=(1-\hat{y}_c^t)^{\alpha \log|\mc{C}^{1:t}|} \log(\hat{y}_c^t), c \in \mc{C}^{t}, \alpha>0,
\end{equation}
where $L_+$ is the positive loss part, $\alpha\log|\mc{C}^{1:t}|$ is an adaptive exponential decay factor related to the class number $|\mc{C}^{1:t}|$ and $\alpha$ is a hyperparameter. 
As the number of learned classes increases, the absence of negative labels becomes more severe, exacerbating the positive-negative imbalance and leading to more false positive errors. The exponential decay factor adaptively controls the intensity of down-weighting, with $\log(\cdot)$ smoothing parameter adjustments.

% Due to the continuously increasing number of learned classes, the absence of negative labels becomes more serious, exacerbating the positive-negative imbalance and resulting in a higher occurrence of false positive errors. The exponential decay factor controls the intensity of down-weighting adaptively. We use $\log(\cdot)$ to make parameter adjustments smoother.

 We then design  $L_{\text{akd}}$ based on  $L_{\text{kd}}$  
to down-weight the contribution of overconfident old task predictions:
\begin{equation}\label{eq:akd1}
L_{\text{akd}} = \sum_{c \in \mc{C}^{1:t-1}} -\hat{y}_c^{t-1} L'_+ - (1-\hat{y}_c^{t-1})  \log(1-\hat{y}_c^t),
\end{equation}

\begin{equation}\label{eq:akd2}
L'_+=(1-\hat{y}_c^t)^{\alpha \log|\mc{C}^{1:t}|} \log(\hat{y}_c^t), c \in \mc{C}^{1:t-1}, \alpha>0 .
\end{equation}
% In Eq.~\eqref{eq:cls2} and Eq.~\eqref{eq:akd2}, 
% The ``$\alpha \log|\mc{C}^{1:t}|$'' can be referred to as the focusing parameter. 

From a qualitative perspective, when a training image in Figure \ref{fig:figure2} (a) contains only ``person'' and ``dog'' labels, the classical KD method may output high prediction scores for non-existent labels like ``car'' and ``cat''.
 % the previous model $f^{t-1}_{\theta}$ may outputs high prediction scores for non-existent labels like ``car'' and ``cat''. 
 AKD can reduce such false positive errors by down-weighting overconfident predictions, thereby rebalancing in loss.
Figure \ref{fig:figure2} (b) presents a quantitative comparison: 1) it shows the prediction score statistics of KD, CSC \cite{du2024confidence} and AKD. Due to the imbalance, KD and CSC produce numerous erroneous and overconfident predictions. Compared to the baseline KD, our method significantly reduces the  FPR from 31.9\% to 2.7\%. Additionally, in comparison to the SOTA method CSC, our approach also demonstrates significant superiority in FPR (2.7\% vs 19.4\%); 
% Additionally, compared to the SOTA method CSC, our approach shows significant superiority in FPR (2.7\% vs. 19.4\%); 
2)  
AKD also significantly improves the performance on mAP, CF1, and OF1 metrics,  indirectly indicating that our method can more effectively mitigate catastrophic forgetting than the baseline KD.

% \subsection{OR for memory label rebalance} 
\subsection{OR for Label Rebalance} 
% To rebalance the positive-negative imbalance and more effectively mitigate catastrophic forgetting at the label augmentation level, we employ a relabeling method to uncover the missing past and future labels of memory samples, called online relabeling.
% 2) Classic replay method collects partially labeled samples, and when combined with the current task’s data for joint training, {there still exists a bias toward positive labels. }
% As a result, these methods fail to enhance the model's ability to distinguish between positive and negative labels, and cannot adequately suppress catastrophic forgetting.
% As illustrated in Figure \ref{fig:figure3}, due to the PL issue, many future and past labels are missing in the memory samples sampled by the classical replay method in SLCIL, which will exacerbate the positive-negative imbalance during subsequent replay. When incorporating the classical replay method based on AKD, the FPR increased from 2.7\% to 6.7\%. 
To mitigate the imbalance at the label level, we propose an approach based on reservoir sampling strategy \cite{rolnick2019experience} called online relabeling (OR). This method completes the labels of memory samples using both past and current models in an online manner. OR consists of two steps: 
% we propose a relabeling method to uncover the missing past and future labels of memory samples, as illustrated in Figure \ref{fig:figure3}. We propose an approach called online relabeling (OR), which completes the labels of the samples in memory using the past and trained current models online. Online Relabeling (OR) consists of two steps. 
% In the first step, we utilize $f_\theta^t$ as the future model to relabel the classes of the old task data (task $1:t-1$) in memory, while we use $f_\theta^{t-1}$ as the past label to relabel the classes of  task $t$ data sampled into memory. For example, as illustrated in Figure 3, after the completion of training for task 3, we use \( f_3 \) to relabel the missing classes of the memory samples from tasks 1 and 2, and then use \( f_2 \) to relabel the missing labels for the memory samples from task 3, continuing this process until all tasks have been trained. This ensures that all labels in memory are completely annotated.

First, after completing training for task $t$, we utilize the trained current model $f_\theta^t$  to relabel new classes $\mc{C}^t$ for the memory data sampled from old task $1:t-1$. Simultaneously, we employ the past model $f_\theta^{t-1}$  to relabel old classes $\mc{C}^{1:t-1}$ for the memory data sampled from the current task $t$.
% $\mc{C}^{1:t-1}$ for task \( t \) memory data. 
For instance, as shown in Figure \ref{fig:figure3}, after completing training for task $3$, we use $f_\theta^3$ to relabel the missing new classes $\mc{C}^3$ (in blue) for the memory data from tasks  $1:2$. Next, we apply $f_\theta^2$ to relabel the missing old classes $\mc{C}^{1:2}$ (in green) for task 3 memory samples.
This label augmentation process continues until all tasks are trained, ensuring that all labels in memory are fully annotated online.
% and this label augmentation process continues until all tasks are trained. This approach ensures that all labels in memory are fully annotated online.
% as shown in Figure \ref{fig:figure3},  the memory samples of the classical replay method only have labels from the current task. We propose OR to complete the missing labels of the samples. Specifically, 
%  in the label block matrix, the future model relabels the missing new task labels found above the main diagonal, while the past model relabels the missing old task labels located below the main diagonal. For example, when Task 2 training is complete, we use the Task 1 model (the past model) to relabel the old classes of Task 2's memory sample and use the Task 2 model (the future model) to relabel the new classes of Task 1's memory samples.
We relabel the missing labels as positive or negative by setting a threshold $n$. A memory sample $x'$ is input into the comparatively reliable model ($f_\theta^{t-1}$ or $f_\theta^{t}$) used for relabeling. If the model outputs a probability bigger than $n$ for class \( i \), the \( i \)-th label is set to 1 (positive); otherwise, it is set to 0 (negative), formulated as  $y'_i = \mathbb{I}(\hat{y}'_i > n)$, 
where  $\mathbb{I}$ is the indicator function that returns 1 if the condition inside is true and 0 otherwise.
% Note that we only use the model from the last task to label the missing old task labels for the current task, without storing any additional models.
In the second step, when training task $t$, after relabeling in the first step, all memory samples are fully annotated. We then sample $(x', y')$ randomly from the memory buffer. 
Given that the first step introduces many negative labels, we use a loss function similar to  $L_{\text{cls}}$ to down-weight these negative labels, rather than using  $L_{\text{bce}}$. The loss function is formulated as:
% \begin{equation}\label{eq:er1}
% L_{\text{er}} = \sum_{c \in \mc{C}^{1:t-1}} -y'_c^t \log(\hat{y}'_c^t) - (1-y'_c^t) L_-,
% \end{equation}
% \begin{equation}\label{eq:er2}
% L_-=(\hat{y}'_c^t)^{\beta \log|\mc{C}^{1:t}|} \log(1-\hat{y}'_c^t), c \in \mc{C}^{1:t-1}, \beta>0 .
% \end{equation}
\begin{equation}\label{eq:er1}
L_{\text{er}} = \sum_{c \in \mc{C}^{1:t-1}} -y'_c \log(\hat{y}'_c) - (1-y'_c) L_-,
\end{equation}
\begin{equation}\label{eq:er2}
L_-=(\hat{y}'_c)^{\beta \log|\mc{C}^{1:t}|} \log(1-\hat{y}'_c), c \in \mc{C}^{1:t-1}, \beta>0 ,
\end{equation}
where $\beta \log|\mc{C}^{1:t}|$ represents the adaptive exponential decay factor,  ensuring balanced learning of positive and negative labels.  
% We also use the adaptive exponential approach to control the intensity of down-weighting, ultimately achieving balanced learning of positive and negative labels in the model. 
OR completes the label matrix in memory, reducing the FPR from 6.7\% to 2.4\%, mitigating label-level imbalance and catastrophic forgetting.

\begin{table*}[t]
	\centering
	\caption{Class-incremental results on PASCAL VOC dataset. 
 %The Joint is offline trained as the \textbf{upper bound}, and Fine-Tuning is the \textbf{lower bound}.  
 A buffer size of 0 means no replay is applied, same as below.} %The final values are the average of the values running with 5 different random seeds.}
	\resizebox{\linewidth}{!}{
	\begin{tabular}{l|c|c|ccc|c|ccc|c}
 	% \begin{tabular}{l|c|c|l|lll|l|lll}
		\hline
		\toprule
        \multirow{3}{*}{\textbf{Method}}& \multirow{3}{*}{\textbf{Backbone}}  & \multirow{2}{*}{\textbf{Buffer}}
        & \multicolumn{4}{|c|}{\textbf{VOC B4-C2}} & \multicolumn{4}{|c}{\textbf{VOC B5-C3}}  \\
        \cline{4-11}
		&&\multirow{2}{*}{\textbf{Size}} & \multicolumn{3}{c|}{Last Acc}& Avg.Acc & \multicolumn{3}{c|}{Last Acc}&  Avg.Acc  \\
        \cline{4-11}
		% &&& mAP (\%) & CF1 & OF1 & mAP (\%) & mAP (\%) & CF1 & OF1 & mAP (\%) \\
		&&&mAP & CF1 & OF1& mAP  & mAP& CF1 & OF1 &  mAP  \\
        \hline
        Joint& CNN  & \multirow{2}{*}{-} & 92.6 & 86.7  & 89.2 & - & 92.6 & 86.7 & 89.2 & -\\
        
        Fine-Tuning& CNN  &  &22.6  & 14.2 &21.4 & 52.8    & 48.7 & 29.1 & 40.3  & 73.4 \\
        
        \hline
        \hline
        %PODNet \cite{douillard2020podnet}& SL  & \multirow{6}{*}{0}  & 25.6  &7.2 & 14.1 & 43.7& 24.7 & 6.8 & 13.9  &  44.3  \\
        
        % LwF \cite{li2017learning}& CNN  &\multirow{6}{*}{0} &  55.6  & 34.9 & 41.8 & 76.4 & 74.1 & 50.5 &45.6 & 84.8  \\
LwF \cite{li2017learning}& CNN  &\multirow{6}{*}{0} &  50.4  & 32.8 & 30.9 & 73.4 & 74.1 & 50.5 &45.6 & 84.8  \\
        AGCN \cite{10221710}& GCN  && 50.2   &35.5& 33.5 &71.2 &71.6 & 55.2 & 51.1&83.1   \\

        KRT \cite{dong2023knowledge}& CAM  &&43.6  & 13.7 & 31.0 &71.0  &74.6 & 39.3 & 46.6 & 86.2   \\

        {CSC} \cite{du2024confidence}& GCN  &&  73.8  & 51.5 & 43.1 & 81.9 & 81.3 & 63.0 & 56.6 & 87.3  \\

% AKD& CNN  && 73.4  & 47.8 & 52.7 & 85.4&81.6 & 67.0 & 62.9&89.2   \\
% AKD& CNN  && 73.1  & 57.2 & 60.0 & 84.6&81.3 & 71.1 & 70.8&88.6   \\
RebLL (AKD)& CNN  && 73.1  & 57.2 & 60.0 & 84.6&79.4 & 64.6 & 66.9&87.7   \\
		\hline
  
            %iCaRL \cite{rebuffi2017icarl}& CIL  & \multirow{6}{*}{20/class} & 59.7 & 19.3 & 22.8 & 43.8   & 65.5 &22.1  & 25.5 & 55.7  \\
        ER \cite{rolnick2019experience}& CNN  & \multirow{3}{*}{2/class}  & 47.1 & 34.7 & 33.1  &  69.9  & 62.8& 50.6 & 47.8&78.3    \\
PODNet \cite{douillard2020podnet}& CNN & &60.4   &  45.3& 38.5 &71.1 &  70.3  &47.7 & 43.3 &81.4     \\
DER++ \cite{buzzega2020dark}& CNN  &  & 61.6 & 33.4 & 29.9 &  77.0  &68.1 & 53.3 & 51.4&78.0    \\
        % ER  + OR& SL  &  & 54.4 & 44.0 & 42.8 & 72.3  & 70.7 & 53.4 &56.0 &82.1    \\
    
        % AGCN-R \cite{10221710}& GCN  &   &  &  &  &    & &  & &    \\
        % AGCN-R  + OR& ML  &   &  &  &  &    & &  & &    \\
        % KRT-R \cite{dong2023knowledge}& CAM  &  &  &  &  &  &    & &  &     \\
        % KRT-R + OR& ML  &  &  &  &  &    & &  & &    \\

        % {CSC-R} \cite{du2024confidence}& GCN &&74.9  &53.1  &43.2  &82.9    &82.2 &63.4  &57.2 &87.9    \\ 
        % {CSC-R} + OR& ML  &  &  &  &  &    & &  & &    \\
        \hline
            RebLL (AKD + OR)& CNN  &  \multirow{1}{*}{2/class} & \textbf{81.0} & \textbf{71.0} & \textbf{70.3} &\textbf{88.5}   &\textbf{84.0} &\textbf{73.9}  &\textbf{72.5} &\textbf{90.3}    \\
            % Ours& GCN  &  &  &  &  &    & &  & &    \\
% AGCN-R  + Ours& ML  &  \multirow{3}{*}{2/class} &  &  &  &    & &  & &    \\
%         KRT-R + Ours& ML  &  &  &  &  &    & &  & &    \\

%         {CSC-R} + Ours& ML  &  &  &  &  &    & &  & &    \\
%     LwF  + ER & SL  & \multirow{6}{*}{2/class}  &  &  &  &    & &  & &    \\    
%        AKD  + OR & ML  &   &  &  &  &    & &  & &    \\
%         AKD  + OR + neg& ML  &   &  &  &  &    & &  & &    \\
% AGCN-R  + AKD + OR& ML  &   &  &  &  &    & &  & &    \\
%         KRT-R + AKD + OR& ML  &  &  &  &  &    & &  & &    \\

%         {CSC-R} + AKD + OR& ML  &  &  &  &  &    & &  & &    \\
		\bottomrule
	\end{tabular}}
\label{tab:results_1}
\end{table*}

\subsection{Final Loss} 
The final loss of RebLL is formulated as follows:
\begin{equation}\label{eq:final}
L = \lambda_\text{akd} L_{\text{cls}} + (1-\lambda_\text{akd}) L_{\text{akd}} + \lambda_\text{er} L_{\text{er}},
\end{equation}
where balancing parameters $\lambda_\text{akd}$ and $\lambda_\text{er}$ control the strength of distillation loss $L_{\text{akd}}$ and replay loss $L_{\text{er}}$.

\section{Experiments}
% This section presents a detailed quantitative comparison of our approach against the latest SOTA methods, supplemented by extensive ablation studies.
\subsection{Datasets and Experimental Setting}

\textbf{Datasets.} We adopt the experimental framework from \cite{dong2023knowledge,du2024confidence}, utilizing the MS-COCO 2014 \cite{lin2014microsoft} and PASCAL VOC 2007 \cite{everingham2010pascal} datasets to validate the efficacy of our method. MS-COCO, a widely recognized benchmark for multi-label classification, includes 82,081 training images and 40,504 validation images across 80 classes, with an average of 2.9 labels per image. The PASCAL VOC 2007 dataset comprises 20 classes, with 5011 images in the training set and 4952 images in the test set, averaging 2.4 labels per image.

\noindent
\textbf{Evaluation Metrics.}
Similar to \cite{dong2023knowledge,du2024confidence}, we employ the widely used metrics of average accuracy (Avg.Acc) and last accuracy (Last Acc) for evaluating CIL tasks. We also use mean average precision (mAP), per-class F1 score (CF1), and overall F1 score (OF1) to assess the performance. The mAP is further divided into average mAP and last mAP, representing the mean mAP across all tasks and the mAP for the final task, respectively. Notably, in the last task, the entire test set is used to evaluate all classes.

\noindent
\textbf{Experiments Setup.}
Following the conventions of prior CIL research \cite{douillard2020podnet, cha2021ssul, dong2023knowledge, du2024confidence}, we define various MLCIL scenarios using the format \{Bx-Cy\}, where ``x'' denotes the number of trained classes in the base task and ``y'' indicates the number of trained classes in each subsequent incremental task. For the VOC 2007 dataset, we use two challenging scenarios \{B4-C2 and B5-C3\}. Similarly, for the  MS-COCO  dataset, we assess RebLL with two challenging scenarios \{B20-C4 and B0-C5\}. There are more incremental tasks in these challenging scenarios than in \cite{dong2023knowledge}. 
The CIL process adheres to the lexicographical order of class names, as described in \cite{dong2023knowledge,du2024confidence}.

\noindent
\textbf{Implementation Details.}
We adhere to the experimental settings in \cite{dong2023knowledge,du2024confidence} to ensure a fair comparison. The training is conducted with a batch size of 64 for 20 epochs on both the PASCAL VOC and MS-COCO datasets.  Similar to CSC \cite{du2024confidence} and KRT \cite{dong2023knowledge}, we use TResNetM \cite{ridnik2021tresnet}  pre-trained on ImageNet-21k as the feature extractor. However, unlike them, we only use the vanilla CNN as our backbone. 
We optimize the network using the Adam optimizer \cite{kingma2014adam} with a weight decay of 1e-4. We apply a consistent learning rate of 4e-5 across all tasks for the PASCAL VOC dataset. For the MS-COCO dataset, we use an initial learning rate of 3e-5 for the base task and 5e-5 for subsequent tasks.  We employ the same data augmentation methods as detailed in \cite{du2024confidence,dong2023knowledge}.
\begin{table*}[t]
	\centering
	\caption{Class-incremental results on MS-COCO dataset. 
 %The Joint is offline trained as the \textbf{upper bound}, and Fine-Tuning is the \textbf{lower bound}.  
} %The final values are the average of the values running with 5 different random seeds.}
	\resizebox{\linewidth}{!}{
	\begin{tabular}{l|c|c|ccc|c|ccc|c}
 	% \begin{tabular}{l|c|c|l|lll|l|lll}
		\hline
		\toprule
        \multirow{3}{*}{\textbf{Method}}& \multirow{3}{*}{\textbf{Backbone}}  & \multirow{2}{*}{\textbf{Buffer}}
        & \multicolumn{4}{|c|}{\textbf{MS-COCO B20-C4}} & \multicolumn{4}{|c}{\textbf{MS-COCO B0-C5}}  \\
        \cline{4-11}
		&&\multirow{2}{*}{\textbf{Size}} & \multicolumn{3}{c|}{Last Acc}& Avg.Acc & \multicolumn{3}{c|}{Last Acc}&  Avg.Acc  \\
        \cline{4-11}
		% &&& mAP (\%) & CF1 & OF1 & mAP (\%) & mAP (\%) & CF1 & OF1 & mAP (\%) \\
		&&&mAP & CF1 & OF1& mAP  & mAP& CF1 & OF1 &  mAP  \\
        \hline
        Joint& CNN & \multirow{2}{*}{-} & 81.8 & 76.4 & 79.4 & - & 81.8 & 76.4 & 79.4 & -\\
        
        Fine-Tuning& CNN  & & 19.4 & 10.9 & 13.4  & 36.5 &22.5  & 15.0 & 23.6 & 48.1     \\
        
        \hline
        \hline
        %PODNet \cite{douillard2020podnet}& SL  & \multirow{6}{*}{0}  & 25.6  &7.2 & 14.1 & 43.7& 24.7 & 6.8 & 13.9  &  44.3  \\
        
        LwF \cite{li2017learning}& CNN  &\multirow{5}{*}{0} &34.6 &17.3  &31.8 &55.4 &  50.6  & 36.3 & 41.1 & 66.2  \\
        AGCN \cite{10221710}& GCN  && 55.6 & 44.2 & 39.6 & 65.7&  53.0  & 43.2 & 41.1 & 64.4    \\

        KRT \cite{dong2023knowledge}& CAM  && 45.2&17.6  & 33.0&64.0&44.5    &22.6  &37.5  &63.1    \\

        % {CSC} \cite{du2024confidence}& GCN  &&  66.6  &52.2  & 49.3 & 74.6 & 63.9 & 46.4 & 42.9& 72.4   \\
        {CSC} \cite{du2024confidence}& GCN  && 60.6 & 44.5 & 43.0& 69.8&  63.4  &50.7  & 50.1 & 71.1    \\

        % AKD& CNN  && 61.4  & 50.6 & 49.0 &71.5 &58.7 &47.2  &43.7 &68.9   \\
        RebLL (AKD)& CNN  &&60.1 &51.3  &49.2 &69.2 & 63.5  & 53.5 & 51.9 &71.7   \\
        % AKD& GCN  &&   &  &  & & &  & &  \\

		\hline

        \hline
        %iCaRL \cite{rebuffi2017icarl}& CIL  & \multirow{6}{*}{20/class} & 59.7 & 19.3 & 22.8 & 43.8   & 65.5 &22.1  & 25.5 & 55.7  \\
        ER \cite{rolnick2019experience}& CNN  & \multirow{3}{*}{5/class}  &41.9 &32.9  &29.8 &53.0& 40.1 & 32.9 &32.3  & 54.6       \\
PODNet \cite{douillard2020podnet}& CNN  && 58.4   &44.0 & 39.1 &67.7  &58.2   &45.1  & 40.8 & 67.2    \\
DER++ \cite{buzzega2020dark}& CNN  &&  57.3  & 41.4 & 35.5 &65.5 &  57.9 &43.6  & 39.2 & 68.2     \\
        % ER  + OR& SL  &  & 53.7 & 35.0 & 43.5 &  64.7  &50.1 &37.4  &40.8 &58.2    \\
    
        % AGCN-R \cite{10221710}& GCN  &   &  &  &  &    & &  & &    \\
        % AGCN-R  + OR& ML  &   &  &  &  &    & &  & &    \\
        % KRT-R \cite{dong2023knowledge}& CAM  &  &  &  &  &    & &  & &    \\
        % KRT-R + OR& ML  &  &  &  &  &    & &  & &    \\

        % {CSC-R} \cite{du2024confidence}& GCN  &  &  &  &  &    & &  & &    \\
        % {CSC-R} + OR& ML  &  &  &  &  &    & &  & &    \\
        \hline
        % Ours& CNN  &  \multirow{1}{*}{5/class} & \textbf{65.5} & \textbf{56.1} & \textbf{54.0} &\textbf{72.5}& 60.5   & 48.1& 51.4 &69.5    \\
        RebLL (AKD + OR)& CNN  &  \multirow{1}{*}{5/class}
        &  \textbf{62.8}  &\textbf{53.3}& \textbf{53.0} &\textbf{71.2}& \textbf{65.5} & \textbf{56.1} & \textbf{54.0} &\textbf{72.5} \\
        % Ours& GCN  &  &  &  &  &    & &  & &    \\

% LwF  + ER & SL  & \multirow{6}{*}{5/class}  &  &  &  &    & &  & &    \\    
         
%         AKD  + OR & ML  &   & 62.8 & 52.9 & 57.8 & 72.4   & &  & &    \\ 
%         AKD  + OR + neg& ML  &   &64.8  & 54.2 & 58.4 &  73.4  & &  & &    \\
% AGCN-R  + AKD + OR& ML  &   &  &  &  &    & &  & &    \\
%         KRT-R + AKD + OR& ML  &  &  &  &  &    & &  & &    \\

%         {CSC-R} + AKD + OR& ML  &  &  &  &  &    & &  & &    \\
		\bottomrule
	\end{tabular}}
\label{tab:results_2}
\end{table*}

\begin{table*}[t]
	\centering
	\caption{Comparative experimental results under different backbones. 
 %The Joint is offline trained as the \textbf{upper bound}, and Fine-Tuning is the \textbf{lower bound}.  
} %The final values are the average of the values running with 5 different random seeds.}
	\resizebox{0.8\linewidth}{!}{
	\begin{tabular}{l|c|ccc|c|ccc|c}
 	% \begin{tabular}{l|c|c|l|lll|l|lll}
		\hline
		\toprule
        \multirow{3}{*}{\textbf{Method}}& \multirow{3}{*}{\textbf{Backbone}}  
        & \multicolumn{4}{|c|}{\textbf{MS-COCO B0-C5}} & \multicolumn{4}{|c}{\textbf{MS-COCO B20-C4}}  \\
        \cline{3-10}
		& & \multicolumn{3}{c|}{Last Acc}& Avg.Acc & \multicolumn{3}{c|}{Last Acc}&  Avg.Acc  \\
        \cline{3-10}
		% &&& mAP (\%) & CF1 & OF1 & mAP (\%) & mAP (\%) & CF1 & OF1 & mAP (\%) \\
		&&mAP & CF1 & OF1& mAP  & mAP& CF1 & OF1 &  mAP  \\
        \hline

        % {CSC} \cite{du2024confidence}& GCN  &&  66.6  &52.2  & 49.3 & 74.6 & 63.9 & 46.4 & 42.9& 72.4   \\
        {L2P} \cite{wang2022learning}& ViT-B/16  &  61.6  & 47.0 & 40.2 & 67.8 & 62.1 & 47.3 & 39.8 & 66.6    \\
        {CSC} \cite{du2024confidence}& GCN  &  63.4  &50.7  & 50.1 & 71.1 & 60.6 & 44.5 & 43.0& 69.8   \\

        \hline
        RebLL& CNN  & 65.5 & 56.1 & 54.0 &72.5&{62.8}  &{53.3}& {53.0} &{71.2} \\
        RebLL& GCN  & \textbf{71.7} & \textbf{62.7} &\textbf{65.3} &\textbf{76.9}&  \textbf{70.6}  &\textbf{60.1} &\textbf{63.1}  &\textbf{75.5}    \\
		\bottomrule
	\end{tabular}}
\label{tab:results_3}
\end{table*}

% Notably, CSC is the SOTA method. It can be observed that AGCN, KRT, and CSC all utilize powerful network architectures as backbones, while our approach, by rebalancing the inherent positive-negative imbalance in MLCIL, allows us to surpass them using only a vanilla CNN.

\subsection{Overall Performance}
We compare eight representative class-incremental methods, including the regularization-based methods LwF \cite{li2017learning}, AGCN \cite{10221710}, KRT  \cite{dong2023knowledge} and CSC \cite{du2024confidence}, the replay-based methods such as ER \cite{rolnick2019experience}, PODNet \cite{douillard2020podnet} and DER++ \cite{buzzega2020dark}, and the prompt-based method L2P \cite{wang2022learning}.
% Following  KRT and CSC, we select Fine-Tuning and Joint as the lower and upper bounds for our method, respectively.
For benchmarking, we use Fine-Tuning and Joint as the lower and upper bounds for our method, respectively, as done by KRT and CSC.
Notably, AGCN, KRT, and CSC are currently the best-performing MLCIL methods, utilizing the powerful graph convolutional network (GCN) or cross-attention mechanism (CAM) as the backbone, with CSC being the current SOTA method. 
% These MLCIL methods all overlook the positive-negative imbalance problem. To rebalance the inherent imbalance in MLCIL, we propose a RebLL framework that comprises AKD and OR. Next, we will demonstrate their effects.
These MLCIL methods, however, do not address the positive-negative imbalance issue. To address this inherent imbalance in MLCIL, we propose the RebLL framework, which includes AKD and OR. Next, we will demonstrate their effects.

\noindent
\textbf{PASCAL VOC.} 
The results for PASCAL VOC in \{B4-C2 and B5-C3\} are presented in Table \ref{tab:results_1}.
 The {RebLL} outperforms other representative methods in all scenarios on PASCAL VOC. We have the following observations: 1) When the buffer size is set to 0, even with a vanilla CNN as the backbone and using AKD alone (without OR),  we outperform the GCN-based SOTA method CSC on many metrics. Specifically, in \{B4-C2\}, we achieve an improvement of \textbf{5.7\%} (51.5\%$\rightarrow$57.2\%) in CF1, \textbf{16.9\%} (43.1\%$\rightarrow$60.0\%) in OF1, and \textbf{2.7\%} (81.9\%$\rightarrow$84.6\%) in average mAP; 2) When the buffer size is configured to 2/class, RebLL {(AKD+OR)} with a vanilla CNN demonstrates significant superiority over all other approaches across every scenario and metric. In \{B4-C2\}, it improves by \textbf{7.2\%} (73.8\%$\rightarrow$81.0\%) in final mAP, \textbf{19.5\%} (51.5\%$\rightarrow$71.0\%) in CF1, \textbf{27.2\%} (43.1\%$\rightarrow$70.3\%) in OF1, and \textbf{6.6\%} (81.9\%$\rightarrow$88.5\%) in average mAP.
% Notably, this significant improvement is primarily due to our method, {AKD and OR}, while the addition of buffer plays a minimal role. Subsequent ablation experiments will provide evidence to support this claim. These similar observations can also be found in the \{B5-C3\} scenario.
Notably, this significant improvement is primarily attributed to the AKD and OR components of our method, with the addition of buffer playing a minimal role. Subsequent ablation experiments will further support this claim. Similar trends are observed in the \{B5-C3\} scenario.

\noindent
\textbf{MS-COCO.} 
The results for MS-COCO in \{B20-C4 and B0-C5\} are presented in Table \ref{tab:results_2}. Similar to the results on PASCAL VOC, RebLL with a vanilla CNN as backbone surpasses the SOTA in all scenarios. For example, in \{B20-C4\}, it improves by \textbf{2.2\%} (60.6\%$\rightarrow$62.8\%) in final mAP, \textbf{8.8\%} (44.5\%$\rightarrow$53.3\%) in CF1, \textbf{10.0\%} (43.0\%$\rightarrow$53.0\%) in OF1, and \textbf{1.4\%} (69.8\%$\rightarrow$71.2\%) in average mAP.

\noindent
\textbf{Comparison of Different Backbones.}
Table \ref{tab:results_3} illustrates the comparison between RebLL, CSC and L2P \cite{wang2022learning}. 
 While L2P utilizes a pre-trained ViT-B/16, CSC employs a GCN backbone, and our method uses a vanilla CNN.
 Remarkably, RebLL achieves SOTA results even with a vanilla CNN.
Furthermore, when we switch to the same GCN backbone as CSC, our performance surpasses that of CSC by a considerable margin in both \{B0-C5 and B20-C4\} of MS-COCO. For example, in \{B0-C5\}, we improved by \textbf{8.3\%} (63.4\%$\rightarrow$71.7\%) in final mAP, \textbf{12.0\%} (50.7\%$\rightarrow$62.7\%) in CF1, \textbf{15.2\%} (50.1\%$\rightarrow$65.3\%) in OF1 and \textbf{5.8\%} (71.1\%$\rightarrow$76.9\%) in average mAP.

\begin{figure}
    \centering
    \includegraphics[width=\linewidth]{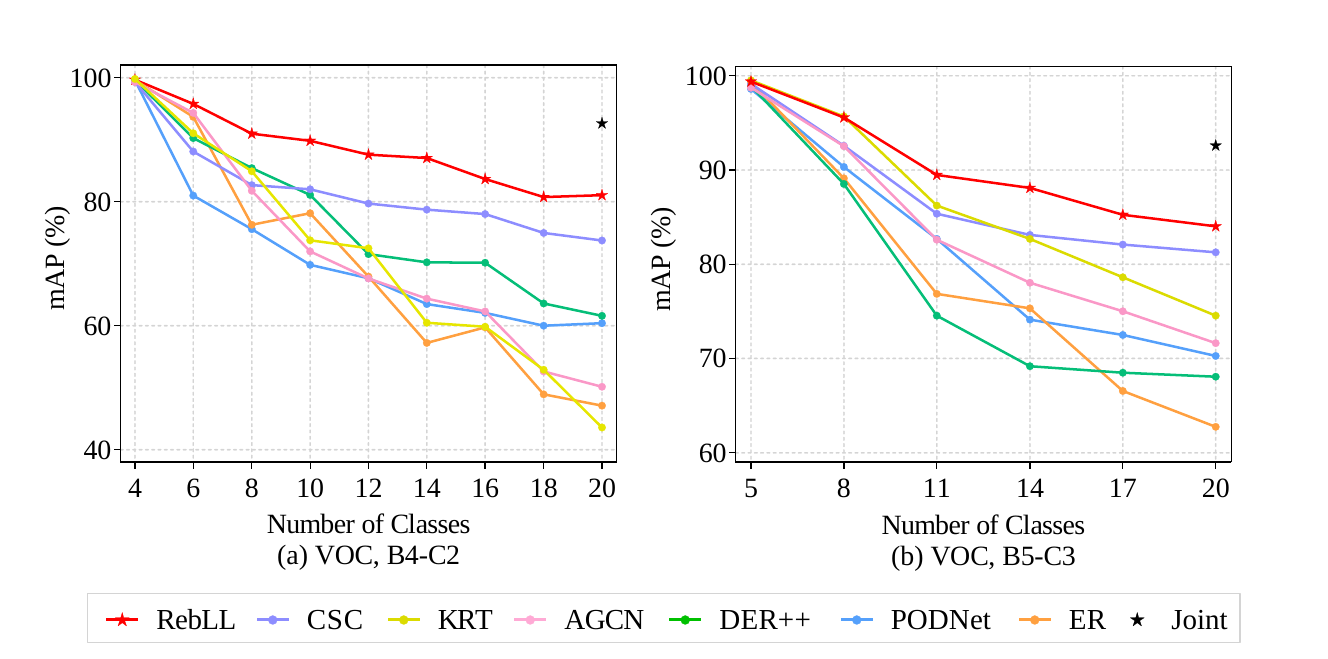}
    \caption{Incremental results on PASCAL VOC in challenging scenarios. There are more tasks in these scenarios.} 
    \label{fig:figure4}
\end{figure}

The incremental results for the challenging scenarios \{B4-C2 and B5-C3\} of PASCAL VOC are shown in Figure \ref{fig:figure4}. These mAP curves illustrate the substantial superiority of RebLL throughout the CIL process.
Additionally, RebLL closely aligns with the upper bound (Joint), showcasing its effectiveness even in long-term class-incremental settings.
\begin{table}[t]
	\centering
	\caption{Quantitative ablation studies for variants of AKD.
} 
	\resizebox{0.85\linewidth}{!}{
	\begin{tabular}{c|ccc|c}

		\hline
		\toprule
\multirow{2}{*}{Method}
        % Fine-Tuning&KD&AKD& AKD
    
		& \multicolumn{3}{c|}{Last Acc}& Avg.Acc \\
  \cline{2-5}

&mAP & CF1 & OF1& mAP   \\
        \hline
       
Fine-Tuning ($L_\text{bce}$)&22.6&14.2&21.4&52.8\\

AKD ($L_\text{cls}$)&36.3&23.3&31.6&56.5\\
KD ($L_\text{bce}+L_\text{kd}$)& 58.2  & 35.3 & 31.0 & 77.5\\
AKD ($L_\text{cls}+L_\text{akd}$) &\textbf{73.1}&\textbf{57.2}&\textbf{60.0}&\textbf{84.6}\\

		\bottomrule
	\end{tabular}}
\label{tab:results_4}
\end{table}

\begin{table}[t]
	\centering
	\caption{Quantitative ablation studies for variants of OR.
} 
	\resizebox{0.85\linewidth}{!}{
	\begin{tabular}{c|ccc|c}

		\hline
		\toprule
\multirow{2}{*}{Method}
        % Fine-Tuning&KD&AKD& AKD
    
		& \multicolumn{3}{c|}{Last Acc}& Avg.Acc \\
  \cline{2-5}

&mAP & CF1 & OF1& mAP   \\
        \hline

AKD ($L_\text{cls}+L_\text{akd}$) &73.1&57.2&60.0&84.6\\
\hline
+ Replay ($L_\text{bce}$)&74.2&63.8&63.9&85.5\\
+ OR ($L_\text{bce}$)&77.3&67.4&68.3&87.2\\
+ OR ($L_\text{er}$)&\textbf{81.0}&\textbf{71.0}&\textbf{70.3}&\textbf{88.5}\\

		\bottomrule
	\end{tabular}}
\label{tab:results_5}
\end{table}

% \begin{table}[t]
% 	\centering
% 	\caption{Quantitative ablation studies for variants of OR.
%  %The Joint is offline trained as the \textbf{upper bound}, and Fine-Tuning is the \textbf{lower bound}.  
% } %The final values are the average of the values running with 5 different random seeds.}
% 	\resizebox{\linewidth}{!}{
% 	\begin{tabular}{c|ccc|ccc|c}
%  	% \begin{tabular}{l|c|c|l|lll|l|lll}
% 		\hline
% 		\toprule
%         % Baseline&Baseline&AKD& AKD& {Replay}& {Replay+OR} & {Replay+OR}
%          AKD& {Replay}& {Replay+OR} & {Replay+OR} 
    
% 		& \multicolumn{3}{c|}{Last Acc}& Avg.Acc \\
%   \cline{5-8}
% 		% &&& mAP (\%) & CF1 & OF1 & mAP (\%) & mAP (\%) & CF1 & OF1 & mAP (\%) \\
% 		($L_\text{cls}+L_\text{akd}$)&($L_\text{bce}$)&($L_\text{bce}$)&($L_\text{er}$)&mAP & CF1 & OF1& mAP   \\
%         \hline

% $\checkmark$&&&&73.1&57.2&60.0&84.6\\
% $\checkmark$&$\checkmark$&&&74.2&63.8&63.9&85.5\\
% $\checkmark$&&$\checkmark$&&77.3&67.4&68.3&87.2\\
% $\checkmark$&&&$\checkmark$&\textbf{81.0}&\textbf{71.0}&\textbf{70.3}&\textbf{88.5}\\

%         % {CSC} \cite{du2024confidence}& GCN  &&  66.6  &52.2  & 49.3 & 74.6 & 63.9 & 46.4 & 42.9& 72.4   \\
%         % {L2P} \cite{wang2022learning}& ViT-B/16  &    &  &  &  &  &  & &    \\
%         % {CSC} \cite{du2024confidence}& GCN  &  63.4  &50.7  & 50.1 & 74.1 & 60.6 & 44.5 & 43.0& 71.0   \\

%         % \hline
%         % Ours& CNN  & 64.8 & 54.2 & 58.4 &73.4& 60.5   & 48.1& 51.4 &69.5    \\
%         % Ours& GCN  &  &  &  &&    & &  &    \\
% 		\bottomrule
% 	\end{tabular}}
% \label{tab:results_5}
% \end{table}

\subsection{Ablation Study}
% We present a comprehensive analysis of our contributions.
\textbf{Effectiveness of AKD.} In Table \ref{tab:results_4},  we use Fine-Tuning ($L_\text{bce}$) and KD ($L_\text{bce} + L_\text{kd}$) as baselines to validate the effectiveness of AKD in scenario \{B4-C2\} of PASCAL VOC. AKD ($L_\text{cls}$) and AKD ($L_\text{cls} + L_\text{akd}$) represent the respective improvements made to Fine-Tuning and KD. 
1) AKD (\(L_{\text{cls}}\)) outperforms Fine-Tuning, highlighting the effectiveness of emphasizing the contribution of negative labels.
 2) From the KD to the AKD ($L_\text{cls} + L_\text{akd}$), the final mAP, CF1 and OF1 are improved by \textbf{14.9\%} (58.2\%$\rightarrow$73.1\%), \textbf{21.9\%} (35.3\%$\rightarrow$57.2\%) and \textbf{29.0\%} (31.0\%$\rightarrow$60.0\%). This indicates the effectiveness of AKD in down-weighting overconfident old task predictions. These observations demonstrate that AKD can perform significantly better by calibrating asymmetric contributions of the positive and negative loss parts to the optimization objectives.

\begin{table}
	\centering
	\caption{The comparison of fixed and adaptive exponential decay factor in scenario \{B4-C2\} of PASCAL VOC.
 %The Joint is offline trained as the \textbf{upper bound}, and Fine-Tuning is the \textbf{lower bound}.  
} %The final values are the average of the values running with 5 different random seeds.}
	\resizebox{\linewidth}{!}{
	\begin{tabular}{c|ccccc}
 	% \begin{tabular}{l|c|c|l|lll|l|lll}
		\hline
		\toprule
        % Baseline&Baseline&AKD& AKD& {Replay}& {Replay+OR} & {Replay+OR}
        decay factor&mAP~$\uparrow$&CF1~$\uparrow$&OF1~$\uparrow$&Avg.mAP~$\uparrow$&FPR~$\downarrow$\\
        \cline{1-6}
$\alpha ~(\alpha=2.0)$&72.0&52.8&57.9&83.5&7.2\\
        $\alpha \log|\mc{C}^{1:t}|  ~(\alpha=1.2)$&\textbf{73.1}&\textbf{57.2}&\textbf{60.0}&\textbf{84.6}&\textbf{2.7}\\

		\bottomrule
	\end{tabular}}
\label{tab:results_6}
\end{table} 
 % This demonstrates the effectiveness of the two componentsa of our proposed AKD.

\noindent
\textbf{Effectiveness of OR.} As shown in Table \ref{tab:results_5}, we conduct ablation of OR based on AKD ($L_\text{cls}+L_\text{akd}$) in scenario \{B4-C2\} of PASCAL VOC. The improvement is minimal when adding the baseline Replay ($L_\text{bce}$) on top of AKD.
OR (\(L_\text{bce}\)) represents the first step of online relabeling, while  OR (\(L_\text{er}\)) means combining the first step with the second step. AKD ($L_\text{cls}+L_\text{akd}$) + OR ($L_\text{er}$) constitute our RebLL framework.
We observe that OR can effectively mitigate label-level imbalance and catastrophic forgetting, thereby enhancing model performance.

% We can observe that OR effectively discovers potential labels and rebalances label-level imbalance by continuously restoring the distribution of labels.
% We can observe that both the first and second steps of OR significantly enhance the effectiveness of the replay method.
% We can see that both the first step Replay + OR ($L_{bce}$) and second step  Replay + OR ($L_{er}$) of OR contribute to improvements in the model, demonstrating the effectiveness of OR.

\begin{figure}[t]
    \centering
    \includegraphics[width=\linewidth]{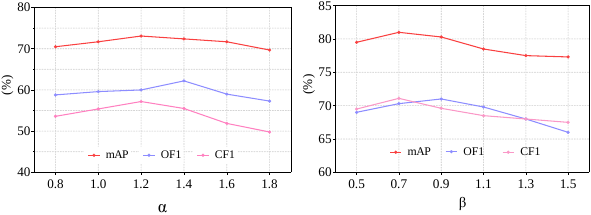}
    \caption{Analysis of $\alpha$ and $\beta$ for exponential decay factor.} 
    \label{fig:figure5}
\end{figure}

\noindent
\textbf{Hyperparameter Analysis.} Figure \ref{fig:figure5} presents an analysis of hyperparameters $\alpha$ and $\beta$  in \{B4-C2\} of PASCAL VOC. When discussing $\alpha$ without including the OR method, AKD reaches its optimum when $\alpha$ is set to 1.2. When OR is incorporated, 
the best performance is achieved with $\beta$ set to 0.7.
% When  $\alpha$ is set to 1.2 and $\beta$ is set to 0.7, the overall performance of the model is superior. 
Table \ref{tab:results_6} compares fixed and adaptive exponential decay factor at the optimal \(\alpha\). 
% As the number of learned classes increases, false positive errors become more frequent. 
% A fixed decay factor cannot adapt to the continuously worsening positive-negative imbalance in MLCIL.
The results demonstrate that a fixed decay factor fails to adapt to the evolving positive-negative imbalance in MLCIL, whereas the adaptive factor significantly outperforms it. 
% A fixed FH that is too large can excessively down-weight early in training, while one that is too small may lead to insufficient down-weighting in later stages. 
% Therefore, we chose adaptive FH, which increases down-weighting intensity as the number of learned classes grows. 
% Table \ref{tab:results_6} shows the superiority of the adaptive exponential decay factor.
% \du{Table \ref{tab:results_6} presents a comparison of fixed and adaptive focusing hyperparameters (FH) at the optimal $\alpha$. Due to the continuously increasing number of learned classes, false positive errors becotheme more frequent. A fixed FH that is too large can excessively down-weight early in the training process, while one that is too small may lead to insufficient down-weighting in the later stages of training. Therefore, we opted for adaptive FH, where the down-weighting intensity increases as the number of learned classes grows. Table \ref{tab:results_6} demonstrates the superiority of the adaptive FH.}
The balancing parameters $\lambda_\text{akd}$ and $\lambda_\text{er}$ are set to 0.15 and 0.30, respectively.
\begin{figure}[t]
    \centering
    \includegraphics[width=\linewidth]{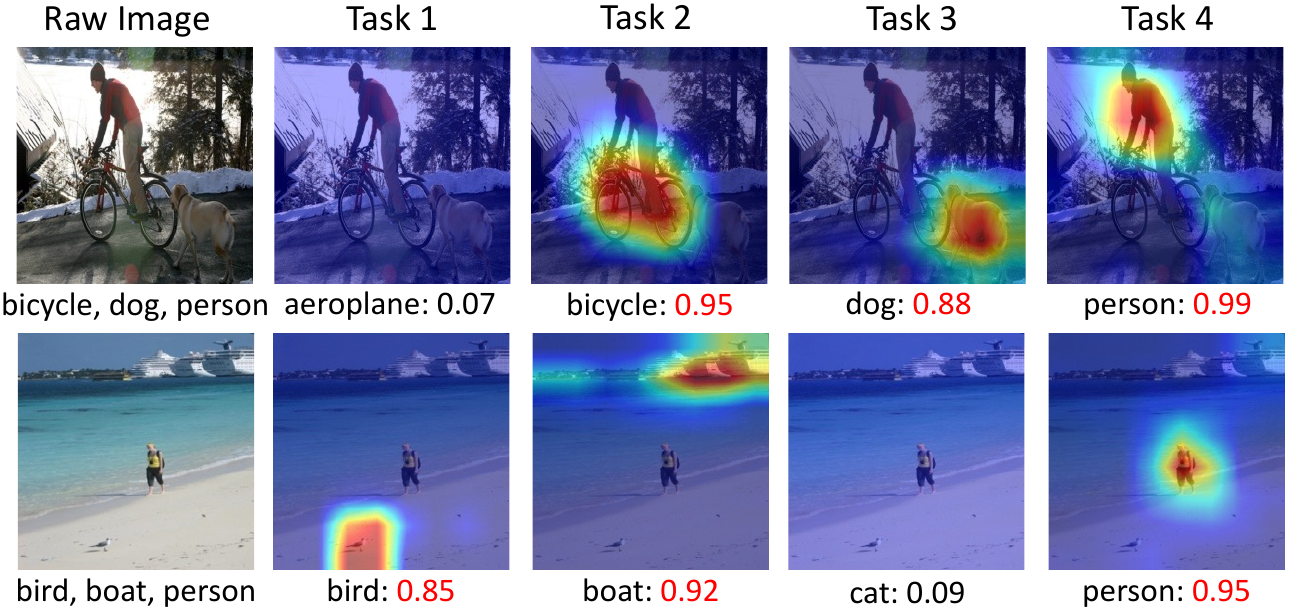}
    \caption{Visualization of RebLL using the final task model.} 
    \label{fig:figure6}
\end{figure}
\subsection{Visualization}

%\ref{fig:figure7} shows several attention maps for four methods. The labels ``bus'', ``cow'', ``horse'' and ``person'' are sequentially trained in different tasks. After training on ``person'', given a test image with ``horse'' and  ``person'', the model outputs attention maps for the four classes. 
As shown in Figure \ref{fig:figure6}, we visualize raw test images alongside their corresponding class activation maps. Given that certain labels are trained sequentially across different tasks, each row presents the class activation map and prediction score for each class using the final task model. We observe that our model effectively highlights label semantic regions, with low activation for non-existent labels. 
For example, in the first row, the activation maps accurately reflect the semantic regions for old task labels such as ``bicycle'', ``dog'' and ``person''. The prediction scores further demonstrate that class-specific representations are highly discriminative. These findings suggest that our model exhibits robust anti-forgetting capabilities, allowing it to precisely capture the semantic regions of labels while demonstrating adequate discernment of positive and negative labels.

\section{Conclusion} 

In this paper, we identify and rebalance the inherent positive-negative imbalance problem in MLCIL under the PL setting. We propose the RebLL framework consisting of two key components: AKD and OR.
AKD rebalances at the loss level by emphasizing the negative label learning and down-weighting the contribution of overconfident predictions. OR is designed for label rebalance, which restores the original class distribution in memory by online relabeling the missing classes. 
Our method is demonstrated to effectively mitigate the positive-negative imbalance, thereby serving as a more efficient anti-forgetting strategy to enhance the MLCIL performance. 
We achieve new state-of-the-art results across various challenging MLCIL scenarios on the PASCAL VOC and MS-COCO datasets, regardless of whether we use a vanilla CNN or a more powerful network architecture as the backbone.

% Limitation. Our superiority over other approaches may be somewhat diminished on simple class-incremental tasks with low forgetting, fewer training tasks and a weaker positive-negative imbalance phenomenon.
\textit{Limitation}. Since we are a rebalancing method. The superiority of our method over other approaches may be less pronounced in simple CIL tasks with minimal forgetting, fewer training tasks, and a weaker positive-negative imbalance phenomenon.


%}

% Using the \centering command instead of \begin{center} ... \end{center} will save space
% Positioning your figure at the top of the page will save space and make the paper more readable
% Using 0.95\columnwidth in conjunction with the

\bibliography{aaai25}

\begin{thebibliography}{37}
\providecommand{\natexlab}[1]{#1}

\bibitem[{Buzzega et~al.(2020)Buzzega, Boschini, Porrello, Abati, and
  Calderara}]{buzzega2020dark}
Buzzega, P.; Boschini, M.; Porrello, A.; Abati, D.; and Calderara, S. 2020.
\newblock Dark experience for general continual learning: a strong, simple
  baseline.
\newblock In \emph{Proceedings of the Advances in Neural Information Processing
  Systems}, 15920--15930.

\bibitem[{Cha et~al.(2023)Cha, Cho, Hwang, Hong, Lee, and
  Moon}]{cha2023rebalancing}
Cha, S.; Cho, S.; Hwang, D.; Hong, S.; Lee, M.; and Moon, T. 2023.
\newblock Rebalancing Batch Normalization for Exemplar-based Class-Incremental
  Learning.
\newblock In \emph{Proceedings of the IEEE/CVF Conference on Computer Vision
  and Pattern Recognition}, 20127--20136.

\bibitem[{Cha et~al.(2021)Cha, Yoo, Moon et~al.}]{cha2021ssul}
Cha, S.; Yoo, Y.; Moon, T.; et~al. 2021.
\newblock SSUL: Semantic segmentation with unknown label for exemplar-based
  class-incremental learning.
\newblock In \emph{Proceedings of the Advances in Neural Information Processing
  Systems}, 10919--10930.

\bibitem[{De~Lange and Tuytelaars(2021)}]{de2021continual}
De~Lange, M.; and Tuytelaars, T. 2021.
\newblock Continual prototype evolution: Learning online from non-stationary
  data streams.
\newblock In \emph{Proceedings of the IEEE/CVF International Conference on
  Computer Vision}, 8250--8259.

\bibitem[{Dong et~al.(2023)Dong, Luo, He, Wei, Cheng, and
  Gong}]{dong2023knowledge}
Dong, S.; Luo, H.; He, Y.; Wei, X.; Cheng, J.; and Gong, Y. 2023.
\newblock Knowledge Restore and Transfer for Multi-Label Class-Incremental
  Learning.
\newblock In \emph{Proceedings of the IEEE/CVF International Conference on
  Computer Vision}, 18711--18720.

\bibitem[{Dosovitskiy et~al.(2021)Dosovitskiy, Beyer, Kolesnikov, Weissenborn,
  Zhai, Unterthiner, Dehghani, Minderer, Heigold, Gelly
  et~al.}]{dosovitskiy2020image}
Dosovitskiy, A.; Beyer, L.; Kolesnikov, A.; Weissenborn, D.; Zhai, X.;
  Unterthiner, T.; Dehghani, M.; Minderer, M.; Heigold, G.; Gelly, S.; et~al.
  2021.
\newblock An image is worth 16x16 words: Transformers for image recognition at
  scale.
\newblock In \emph{Proceedings of the International Conference on Learning
  Representations}.

\bibitem[{Douillard et~al.(2020)Douillard, Cord, Ollion, Robert, and
  Valle}]{douillard2020podnet}
Douillard, A.; Cord, M.; Ollion, C.; Robert, T.; and Valle, E. 2020.
\newblock Podnet: Pooled outputs distillation for small-tasks incremental
  learning.
\newblock In \emph{Proceedings of the European Conference on Computer Vision},
  86--102.

\bibitem[{Douillard et~al.(2022)Douillard, Ram{\'e}, Couairon, and
  Cord}]{douillard2022dytox}
Douillard, A.; Ram{\'e}, A.; Couairon, G.; and Cord, M. 2022.
\newblock Dytox: Transformers for continual learning with dynamic token
  expansion.
\newblock In \emph{Proceedings of the IEEE/CVF Conference on Computer Vision
  and Pattern Recognition}, 9285--9295.

\bibitem[{Du et~al.(2022)Du, Lyu, Hu, Li, Feng, Xu, and Fu}]{9859622}
Du, K.; Lyu, F.; Hu, F.; Li, L.; Feng, W.; Xu, F.; and Fu, Q. 2022.
\newblock AGCN: Augmented Graph Convolutional Network for Lifelong Multi-Label
  Image Recognition.
\newblock In \emph{IEEE International Conference on Multimedia and Expo},
  01--06.

\bibitem[{Du et~al.(2024{\natexlab{a}})Du, Lyu, Li, Hu, Feng, Xu, Xi, and
  Cheng}]{10221710}
Du, K.; Lyu, F.; Li, L.; Hu, F.; Feng, W.; Xu, F.; Xi, X.; and Cheng, H.
  2024{\natexlab{a}}.
\newblock Multi-Label Continual Learning Using Augmented Graph Convolutional
  Network.
\newblock \emph{IEEE Transactions on Multimedia}, 26: 2978--2992.

\bibitem[{Du et~al.(2024{\natexlab{b}})Du, Zhou, Lyu, Li, Lu, and
  Liu}]{du2024confidence}
Du, K.; Zhou, Y.; Lyu, F.; Li, Y.; Lu, C.; and Liu, G. 2024{\natexlab{b}}.
\newblock Confidence Self-Calibration for Multi-Label Class-Incremental
  Learning.
\newblock In \emph{Proceedings of the European Conference on Computer Vision}.

\bibitem[{Everingham et~al.(2010)Everingham, Van~Gool, Williams, Winn, and
  Zisserman}]{everingham2010pascal}
Everingham, M.; Van~Gool, L.; Williams, C.~K.; Winn, J.; and Zisserman, A.
  2010.
\newblock The pascal visual object classes (voc) challenge.
\newblock \emph{International Journal of Computer Vision}.

\bibitem[{Kim, Jeong, and Kim(2020)}]{kim2020imbalanced}
Kim, C.~D.; Jeong, J.; and Kim, G. 2020.
\newblock Imbalanced continual learning with partitioning reservoir sampling.
\newblock In \emph{Proceedings of the European Conference on Computer Vision},
  411--428.

\bibitem[{Kingma and Ba(2015)}]{kingma2014adam}
Kingma, D.~P.; and Ba, J. 2015.
\newblock Adam: A method for stochastic optimization.
\newblock In \emph{Proceedings of the International Conference on Learning
  Representations}.

\bibitem[{Kirkpatrick et~al.(2017)Kirkpatrick, Pascanu, Rabinowitz, Veness,
  Desjardins, Rusu, Milan, Quan, Ramalho, Grabska-Barwinska
  et~al.}]{kirkpatrick2017overcoming}
Kirkpatrick, J.; Pascanu, R.; Rabinowitz, N.; Veness, J.; Desjardins, G.; Rusu,
  A.~A.; Milan, K.; Quan, J.; Ramalho, T.; Grabska-Barwinska, A.; et~al. 2017.
\newblock Overcoming catastrophic forgetting in neural networks.
\newblock \emph{National Academy of Sciences}, 114(13): 3521--3526.

\bibitem[{Li and Hoiem(2017)}]{li2017learning}
Li, Z.; and Hoiem, D. 2017.
\newblock Learning without forgetting.
\newblock \emph{IEEE Transactions on Pattern Analysis and Machine
  Intelligence}, 40(12): 2935--2947.

\bibitem[{Liang and Li(2022)}]{liang2022optimizing}
Liang, Y.-S.; and Li, W.-J. 2022.
\newblock Optimizing Class Distribution in Memory for Multi-Label Online
  Continual Learning.
\newblock \emph{arXiv preprint arXiv:2209.11469}.

\bibitem[{Lin et~al.(2017)Lin, Goyal, Girshick, He, and
  Doll{\'a}r}]{lin2017focal}
Lin, T.-Y.; Goyal, P.; Girshick, R.; He, K.; and Doll{\'a}r, P. 2017.
\newblock Focal loss for dense object detection.
\newblock In \emph{Proceedings of the IEEE/CVF International Conference on
  Computer Vision}, 2980--2988.

\bibitem[{Lin et~al.(2014)Lin, Maire, Belongie, Hays, Perona, Ramanan,
  Doll{\'a}r, and Zitnick}]{lin2014microsoft}
Lin, T.-Y.; Maire, M.; Belongie, S.; Hays, J.; Perona, P.; Ramanan, D.;
  Doll{\'a}r, P.; and Zitnick, C.~L. 2014.
\newblock Microsoft coco: Common objects in context.
\newblock In \emph{Proceedings of the European Conference on Computer Vision},
  740--755.

\bibitem[{Liu et~al.(2023)Liu, Xu, Lv, and Geng}]{liu2023revisiting}
Liu, B.; Xu, N.; Lv, J.; and Geng, X. 2023.
\newblock Revisiting pseudo-label for single-positive multi-label learning.
\newblock In \emph{International Conference on Machine Learning}, 22249--22265.
  PMLR.

\bibitem[{Luo et~al.(2023)Luo, Liu, Schiele, and Sun}]{luo2023class}
Luo, Z.; Liu, Y.; Schiele, B.; and Sun, Q. 2023.
\newblock Class-incremental exemplar compression for class-incremental
  learning.
\newblock In \emph{Proceedings of the IEEE/CVF Conference on Computer Vision
  and Pattern Recognition}, 11371--11380.

\bibitem[{Lyu et~al.(2024)Lyu, Du, Li, Zhao, Zhang, Liu, and
  Wang}]{lyu2024variational}
Lyu, F.; Du, K.; Li, Y.; Zhao, H.; Zhang, Z.; Liu, G.; and Wang, L. 2024.
\newblock Variational Continual Test-Time Adaptation.
\newblock \emph{arXiv preprint arXiv:2402.08182}.

\bibitem[{Lyu et~al.(2023)Lyu, Sun, Shang, Wan, and Feng}]{lyu2023measuring}
Lyu, F.; Sun, Q.; Shang, F.; Wan, L.; and Feng, W. 2023.
\newblock Measuring asymmetric gradient discrepancy in parallel continual
  learning.
\newblock In \emph{Proceedings of the IEEE/CVF International Conference on
  Computer Vision}, 11411--11420.

\bibitem[{Lyu et~al.(2021)Lyu, Wang, Feng, Ye, Hu, and Wang}]{lyu2020multi}
Lyu, F.; Wang, S.; Feng, W.; Ye, Z.; Hu, F.; and Wang, S. 2021.
\newblock Multi-domain multi-task rehearsal for lifelong learning.
\newblock In \emph{Proceedings of the AAAI Conference on Artificial
  Intelligence}, 8819--8827.

\bibitem[{McCloskey and Cohen(1989)}]{mccloskey1989catastrophic}
McCloskey, M.; and Cohen, N.~J. 1989.
\newblock Catastrophic interference in connectionist networks: The sequential
  learning problem.
\newblock In \emph{Psychology of Learning and Motivation}, volume~24, 109--165.

\bibitem[{Mohamed et~al.(2023)Mohamed, Grandhe, Joseph, Khan, and
  Khan}]{mohamed2023d3former}
Mohamed, A.; Grandhe, R.; Joseph, K.; Khan, S.; and Khan, F. 2023.
\newblock D3Former: Debiased Dual Distilled Transformer for Incremental
  Learning.
\newblock In \emph{Proceedings of the IEEE/CVF Conference on Computer Vision
  and Pattern Recognition}, 2420--2429.

\bibitem[{Ridnik et~al.(2021{\natexlab{a}})Ridnik, Ben-Baruch, Zamir, Noy,
  Friedman, Protter, and Zelnik-Manor}]{ridnik2021asymmetric}
Ridnik, T.; Ben-Baruch, E.; Zamir, N.; Noy, A.; Friedman, I.; Protter, M.; and
  Zelnik-Manor, L. 2021{\natexlab{a}}.
\newblock Asymmetric loss for multi-label classification.
\newblock In \emph{Proceedings of the IEEE/CVF International Conference on
  Computer Vision}, 82--91.

\bibitem[{Ridnik et~al.(2021{\natexlab{b}})Ridnik, Lawen, Noy, Ben~Baruch,
  Sharir, and Friedman}]{ridnik2021tresnet}
Ridnik, T.; Lawen, H.; Noy, A.; Ben~Baruch, E.; Sharir, G.; and Friedman, I.
  2021{\natexlab{b}}.
\newblock Tresnet: High performance gpu-dedicated architecture.
\newblock In \emph{Proceedings of the IEEE/CVF Winter Conference on
  Applications of Computer Vision}, 1400--1409.

\bibitem[{Rolnick et~al.(2019)Rolnick, Ahuja, Schwarz, Lillicrap, and
  Wayne}]{rolnick2019experience}
Rolnick, D.; Ahuja, A.; Schwarz, J.; Lillicrap, T.; and Wayne, G. 2019.
\newblock Experience replay for continual learning.
\newblock In \emph{Proceedings of the Advances in Neural Information Processing
  Systems}, 350--360.

\bibitem[{Sun et~al.(2022)Sun, Lyu, Shang, Feng, and Wan}]{sun2022exploring}
Sun, Q.; Lyu, F.; Shang, F.; Feng, W.; and Wan, L. 2022.
\newblock Exploring example influence in continual learning.
\newblock In \emph{Proceedings of the Advances in Neural Information Processing
  Systems}, volume~35, 27075--27086.

\bibitem[{Wang et~al.(2024)Wang, Zhang, Su, and Zhu}]{wang2023comprehensive}
Wang, L.; Zhang, X.; Su, H.; and Zhu, J. 2024.
\newblock A Comprehensive Survey of Continual Learning: Theory, Method and
  Application.
\newblock \emph{IEEE Transactions on Pattern Analysis and Machine
  Intelligence}, 46(8): 5362--5383.

\bibitem[{Wang et~al.(2022)Wang, Zhang, Lee, Zhang, Sun, Ren, Su, Perot, Dy,
  and Pfister}]{wang2022learning}
Wang, Z.; Zhang, Z.; Lee, C.-Y.; Zhang, H.; Sun, R.; Ren, X.; Su, G.; Perot,
  V.; Dy, J.; and Pfister, T. 2022.
\newblock Learning to prompt for continual learning.
\newblock In \emph{Proceedings of the IEEE/CVF Conference on Computer Vision
  and Pattern Recognition}, 139--149.

\bibitem[{Wu et~al.(2021)Wu, Baek, You, and Ma}]{wu2021incremental}
Wu, Z.; Baek, C.; You, C.; and Ma, Y. 2021.
\newblock Incremental learning via rate reduction.
\newblock In \emph{Proceedings of the IEEE/CVF Conference on Computer Vision
  and Pattern Recognition}, 1125--1133.

\bibitem[{Xie et~al.(2024)Xie, Xiao, Liu, Niu, Sugiyama, and
  Huang}]{xie2024class}
Xie, M.-K.; Xiao, J.; Liu, H.-Z.; Niu, G.; Sugiyama, M.; and Huang, S.-J. 2024.
\newblock Class-distribution-aware pseudo-labeling for semi-supervised
  multi-label learning.
\newblock \emph{Advances in Neural Information Processing Systems}, 36.

\bibitem[{Ye and Bors(2022)}]{ye2021lifelong}
Ye, F.; and Bors, A.~G. 2022.
\newblock Lifelong Teacher-Student Network Learning.
\newblock \emph{IEEE Transactions on Pattern Analysis and Machine
  Intelligence}, 44(10): 6280--6296.

\bibitem[{Zhao et~al.(2023)Zhao, Lu, Xu, Cheng, Guo, Niu, and
  Fang}]{zhao2023few}
Zhao, L.; Lu, J.; Xu, Y.; Cheng, Z.; Guo, D.; Niu, Y.; and Fang, X. 2023.
\newblock Few-Shot Class-Incremental Learning via Class-Aware Bilateral
  Distillation.
\newblock In \emph{Proceedings of the IEEE/CVF Conference on Computer Vision
  and Pattern Recognition}, 11838--11847.

\bibitem[{Zhou et~al.(2022)Zhou, Chen, Wang, Chen, and
  Heng}]{zhou2022acknowledging}
Zhou, D.; Chen, P.; Wang, Q.; Chen, G.; and Heng, P.-A. 2022.
\newblock Acknowledging the unknown for multi-label learning with single
  positive labels.
\newblock In \emph{European Conference on Computer Vision}, 423--440. Springer.

\end{thebibliography}

\end{document}